\documentclass[10pt,journal,compsoc]{IEEEtran}

\usepackage[nocompress]{cite}

\usepackage{url}
\usepackage{ragged2e}
\usepackage{epsfig}
\usepackage{graphicx}
\usepackage{amsmath,amssymb} 
\usepackage{mathptmx}      
\usepackage{subfigure}
\usepackage{algorithm}
\usepackage{algorithmicx}
\usepackage{algpseudocode}
\usepackage{graphics}
\usepackage{mathrsfs}
\usepackage{threeparttable}
\usepackage{color}
\usepackage[normalem]{ulem}
\usepackage{multirow}
\usepackage{float}
\usepackage{amsfonts}
\usepackage{bm}
\usepackage{array}
\usepackage[table]{xcolor}
\usepackage{colortbl}
\usepackage{pifont}
\usepackage{diagbox}
\usepackage{rotating}
\usepackage{booktabs}
\usepackage{overpic}
\usepackage{textcomp}
\usepackage{contour}
\usepackage{enumitem}
\usepackage{stfloats}
\usepackage{threeparttable}
\usepackage{makecell}
\usepackage[colorlinks=true,breaklinks=true,urlcolor=magenta]{hyperref}
\usepackage{xspace}

\def\ie{\emph{i.e.}}
\def\eg{\emph{e.g.}}

\newcommand{\Bard}{\textsc{Bard}\xspace}

\newcommand{\figref}[1]{Fig.~\ref{#1}}

\newcommand{\secref}[1]{\S\ref{#1}}

\def\ie{\textit{i.e.}}
\def\eg{\textit{e.g.}}

\graphicspath{{./Imgs/}}
\DeclareGraphicsExtensions{.jpg,.pdf,.png}

\RequirePackage{silence}
\begin{document}

\title{\huge How Good is Google Bard's Visual Understanding? \\ An Empirical Study on Open Challenges}

\author{
Haotong Qin$^\dagger$,~
Ge-Peng Ji$^\dagger$,~
Salman Khan,~
Deng-Ping Fan$^*$,~
Fahad Shahbaz Khan,~
Luc Van Gool\\
\IEEEcompsocitemizethanks{
\IEEEcompsocthanksitem Haotong Qin, Deng-Ping Fan, and Luc Van Gool are with the Computer Vision Lab (CVL), ETH Zurich, Zurich, Switzerland.
\IEEEcompsocthanksitem Ge-Peng Ji is with the College of Engineering, Computing \& Cybernetics, ANU, Canberra, Australia.
\IEEEcompsocthanksitem Salman Khan and Fahad Shahbaz Khan are with Mohamed bin Zayed University of Artificial Intelligence, Abu Dhabi, UAE.
\IEEEcompsocthanksitem $\dagger$: Equal contribution; $*$: Corresponding author (dengpfan@gmail.com).
}
}


\IEEEtitleabstractindextext{%
\begin{abstract} \justifying
Google's \Bard has emerged as a formidable competitor to OpenAI's ChatGPT in the field of conversational AI. Notably, \Bard has recently been updated to handle visual inputs alongside text prompts during conversations. Given \Bard's impressive track record in handling textual inputs, we explore its capabilities in understanding and interpreting visual data (images) conditioned by text questions. This exploration holds the potential to unveil new insights and challenges for \Bard and other forthcoming multi-modal Generative models, especially in addressing complex computer vision problems that demand accurate  visual and language understanding. Specifically, in this study, we focus on 15 diverse task scenarios encompassing regular, camouflaged, medical, under-water and remote sensing data to comprehensively evaluate \Bard's performance. Our primary finding indicates that \Bard still struggles in these vision scenarios, highlighting the significant gap in vision-based understanding that needs 
to be bridged in future developments. We expect that this empirical study will prove valuable in advancing future models, leading to enhanced capabilities in comprehending and interpreting fine-grained visual data. 
Our project is released on \url{https://github.com/htqin/GoogleBard-VisUnderstand}.
\end{abstract}

}

\maketitle

\IEEEdisplaynontitleabstractindextext

\IEEEpeerreviewmaketitle

\begin{figure*}[ht!]
  \centering
  \vspace{-10pt}
  \includegraphics[width=\linewidth]{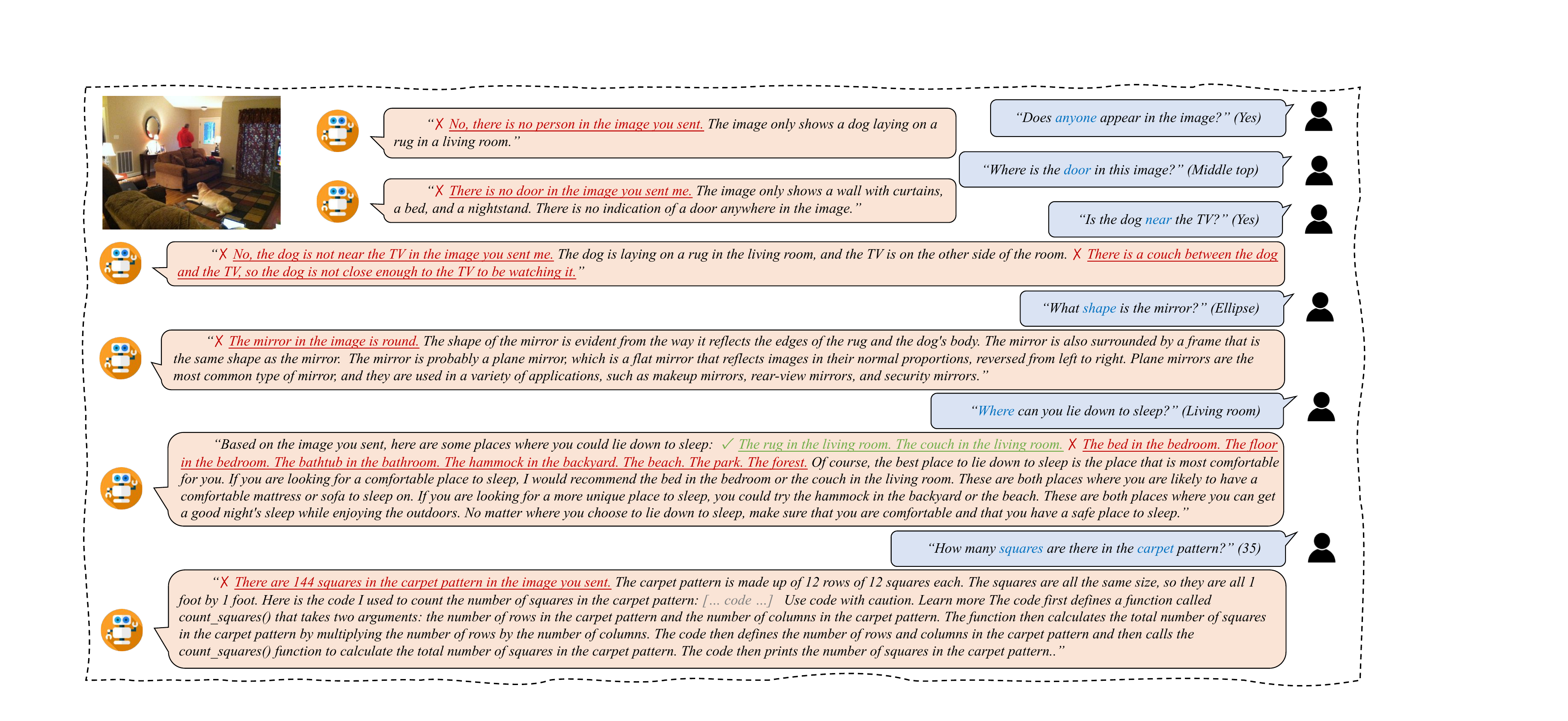}
  \vspace{-20pt}
  \caption{Several examples of multi-modal interactive sessions using Google's \Bard, wherein the AI system responds to the user's question based on images sourced from the Microsoft COCO dataset~\cite{lin2014microsoft}. In the following figures, the expected standard answers are provided in parentheses, except where otherwise stated. Please refer to~\secref{sec:scenario_1},~\secref{sec:scenario_2},~\secref{sec:scenario_3},~\secref{sec:scenario_4},~\secref{sec:scenario_5}, and~\secref{sec:camofish5k}, for further details.}
  \label{fig:microsoft_coco}
\end{figure*}

\begin{figure*}[ht!]
  \centering
  \vspace{-5pt}
  \includegraphics[width=\linewidth]{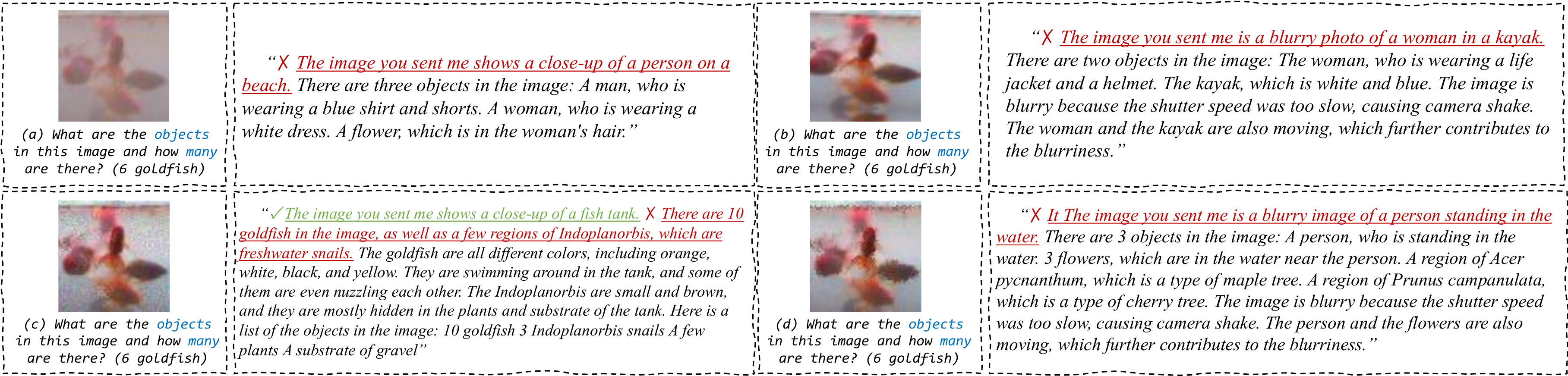}
  \vspace{-20pt}
  \caption{Several examples of multi-modal interactive sessions using Google's \Bard, wherein the AI system responds to the user's question based on images sourced from the Tiny-ImageNet-C dataset~\cite{hendrycks2019benchmarking}. Please refer to~\secref{sec:adversarial} for further details.}
  \label{fig:adversarial}
\end{figure*}

\IEEEraisesectionheading{\section{Introduction}\label{sec:introduction}}

\IEEEPARstart{B}{ard}\footnote{\url{https://bard.google.com}}, Google's AI chatbot, based on LaMDA~\cite{thoppilan2022lamda} and later PaLM~\cite{chowdhery2022palm} models, was launched with moderate success in March 2023 before expanding globally in May. It’s a generative AI that accepts prompts and performs text-based tasks like providing answers, and summaries, and creating various forms of text content.
On 13 July 2023, Google \Bard announced a major update\footnote{\scriptsize\url{https://blog.google/products/bard/google-bard-new-features-update-july-2023/}} which allowed providing images as inputs together with textual prompts. It was claimed that \Bard\ can analyze visual content and provide a description (\eg, image captions) or answer questions using visual information. Notably, although other models such as GPT4~\cite{openai2023gpt4} have claimed to have capabilities to accept and understand visual inputs as prompts, they are not publicly accessible for experimentation. Therefore, access to \Bard\ provides a first opportunity for the computer vision community to assess its soundness and robustness toward understanding existing strengths and limitations. In this empirical study, our goal is to analyze the capability of \Bard\ towards some of the long-standing problems of computer vision in image comprehension. 

Our study identifies several interesting scenarios based on computer vision problems for the qualitative evaluation of \Bard. Since API-based access to \Bard\ is still not available, our evaluations do not comprise of quantitative results on large-scale benchmarks. Instead, our goal is to identify a number of insightful scenarios and corresponding visual-textual prompts that serves the purpose of evaluating not only the visual understanding capabilities of \Bard\ but future large multimodal models such as GPT4 as well. Our motivation to particularly focus on \Bard is its top performance among all open and closed-source multimodal conversational models (including Bing-Chat rolled out on 18 July 2023 \cite{Bing-chat}) as demonstrated via LLaVA-Bench \cite{LLaVA-Bench}.


\section{Empirical Experiments}\label{sec:experiment}

To assess \Bard's capabilities, such as visual perception and contextual understanding, conditioned on the given text prompts, we designed a range of vision-language task scenarios. Subsequently, we delve into several illustrative examples drawn from these empirical studies, encompassing a total of 15 visual question-answering (VQA) scenarios involving tasks such as object detection and localization, analyzing object attributes, count, affordances, and fine-grained recognition in natural images. We also experiment with challenging cases such as identifying camouflaged objects and diverse domains such as medical, underwater, and remote sensing images. We explain the scenarios below.

\subsection{Scenario~\#1 -- Object attributes}\label{sec:scenario_1}

Understanding the properties and characteristics of objects within an image is a fundamental task in VQA. For instance, in the fourth question of \figref{fig:microsoft_coco}, when presented with the query, `\texttt{What shape is the mirror?}'
\Bard\ fails to understand the shape-related attributes of the object mirror and also hallucinates about the reflection appearing in it. This suggests that \Bard\ appears to have challenges in identifying attributes that necessitate a deep understanding of each object and its properties.

\subsection{Scenario~\#2 -- Object presence}\label{sec:scenario_2}
This evaluates \Bard's ability to identify a specific object conditioned by a provided text description. As evidenced by the first question in \figref{fig:microsoft_coco}, \Bard~fails to answer the question correctly, `\texttt{Does anyone appear in the image?}', and provides an incorrect response, `\texttt{There is no person in the image}'. This suggests that \Bard's basic understanding of visual content remains limited. We further note that \Bard is currently tailored for images without any humans and deletes any visual inputs containing human faces or persons.

\subsection{Scenario~\#3 -- Object location}\label{sec:scenario_3}

This task scenario examines \Bard's ability to locate and understand objects within an image. For example, reference the second question in \figref{fig:microsoft_coco}, inquiring, `\texttt{Where is the door in this image?}' However, \Bard~is unable to identify the door in the provided image, replying, `\texttt{There is no door in the image you sent me.}' Thus, this suggests that \Bard's localization ability of visual context can be further enhanced.

\subsection{Scenario~\#4 -- Relationship reasoning}\label{sec:scenario_4}

This scenario evaluates \Bard's ability to understand complex inter-object dynamics within an image, scrutinizing its understanding of spatial and semantic relationships. As depicted in the final question in \figref{fig:microsoft_coco}, we query \Bard, `\texttt{Is the dog near the TV?}' However, it fails to provide correct feedback, replying, `\texttt{No, the dog is not near the TV in the image you sent me.}' Therefore, this indicates that there is room to improve \Bard's ability in reasoning relationships.

\subsection{Scenario~\#5 -- Affordance}\label{sec:scenario_5}

The affordance test is used to validate \Bard's understanding ability in potential uses or actions that an object can offer or support. This delves into the model's understanding of functionality beyond mere object recognition. An exemplar study is the fifth question in \figref{fig:microsoft_coco}, `\texttt{Where can you lie down to sleep?}'. Interestingly, \Bard\ can provide two plausible responses (\ie, rug and couch), but fails to do so consistently for other options, such as some absent items: `\texttt{The bed in the bedroom.}' and `\texttt{The beach. The park. The forest.}' This hallucination in the outputs implies that \Bard~still needs to better capture visual semantics strictly based on the text guidance and more effectively associate these semantics with recognized objects in a scene.





\subsection{Scenario~\#6 -- Adversarial sample}\label{sec:adversarial}

Adversarial samples refer to carefully crafted inputs intended to confuse or mislead AI models. Such inputs remain perceptually understandable to humans due to minor changes in pixel values, however, become difficult to interpret for an AI model. Exploring \Bard's response to these samples in a black-box setting is vital as it provides a measure of the model's robustness and stability. As depicted in \figref{fig:adversarial}, we present four groups of image-text pairs where the image is showing a number of fish. All outputs from \Bard demonstrate that it fails to understand adversarial samples.

\subsection{Scenario~\#7 -- Rainy conditions}\label{sec:rain}

\begin{figure*}[ht!]
  \centering
  \vspace{-5pt}
  \includegraphics[width=\linewidth]{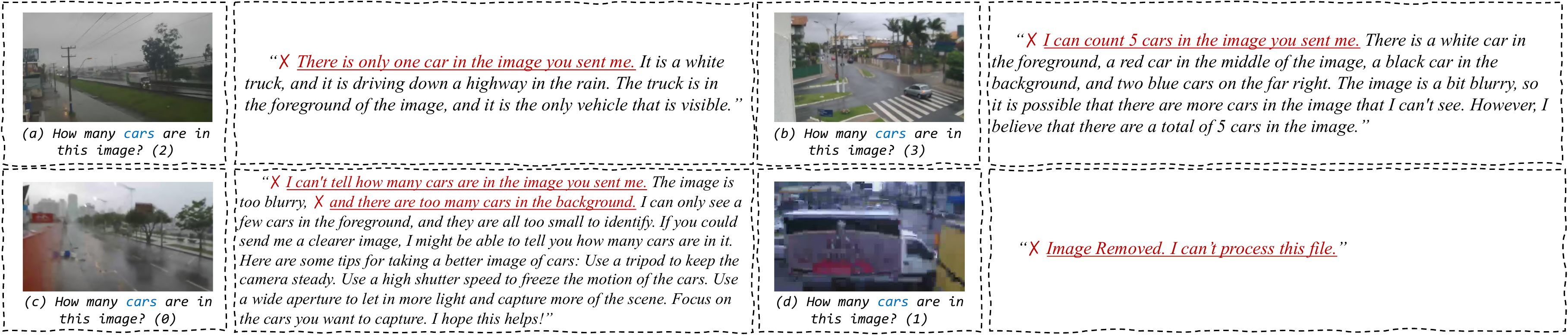}
  \vspace{-20pt}
  \caption{Several examples of multi-modal interactive sessions using Google's \Bard, wherein the AI system responds to the user's question based on images sourced from the MPID dataset~\cite{Li_2019_CVPR}. Please refer to~\secref{sec:rain} for further details.}
  \label{fig:rain}
\end{figure*}

Different weather conditions play a crucial role in assessing a model's ability to handle unanticipated situations and domain shifts, an important requisite for real-world applicability. Consequently, we selected four groups of samples to test \Bard~under natural scenarios with unique weather conditions, such as rain as depicted in \figref{fig:rain}. The results indicate that \Bard~does not perform well when the image features rainy conditions.

\begin{figure*}[ht!]
  \centering
  \vspace{-5pt}
  \includegraphics[width=\linewidth]{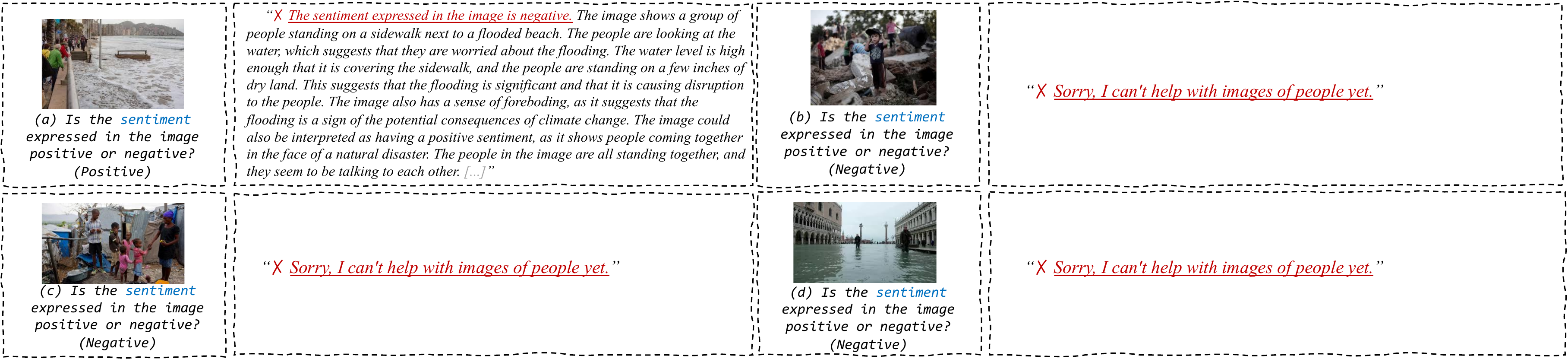}
  \vspace{-20pt}
  \caption{Several examples of multi-modal interactive sessions using Google's \Bard, wherein the AI system responds to the user's question based on images sourced from the Image Sentiment dataset~\cite{hassan2020visual}. Please refer to~\secref{sec:sentiment} for further details.}
  \label{fig:sentiment}
\end{figure*}

\subsection{Scenario~\#8 -- Sentiment understanding}\label{sec:sentiment}

This scenario evaluates \Bard's capability to understand the sentiment or emotional context of a scene in the image, thus examining its comprehension of more abstract, non-tangible aspects of visual data. As demonstrated in \figref{fig:sentiment}, when we query \Bard, `\texttt{Is the sentiment expressed in the image positive or negative?}', it replies an incorrect response, `\texttt{ The sentiment expressed in the image is negative.}'

\subsection{Scenario~\#9 -- Fine-grained recognition}
\label{sec:fgvc}
\begin{figure*}[ht!]
  \centering
  \vspace{-5pt}
  \includegraphics[width=\linewidth]{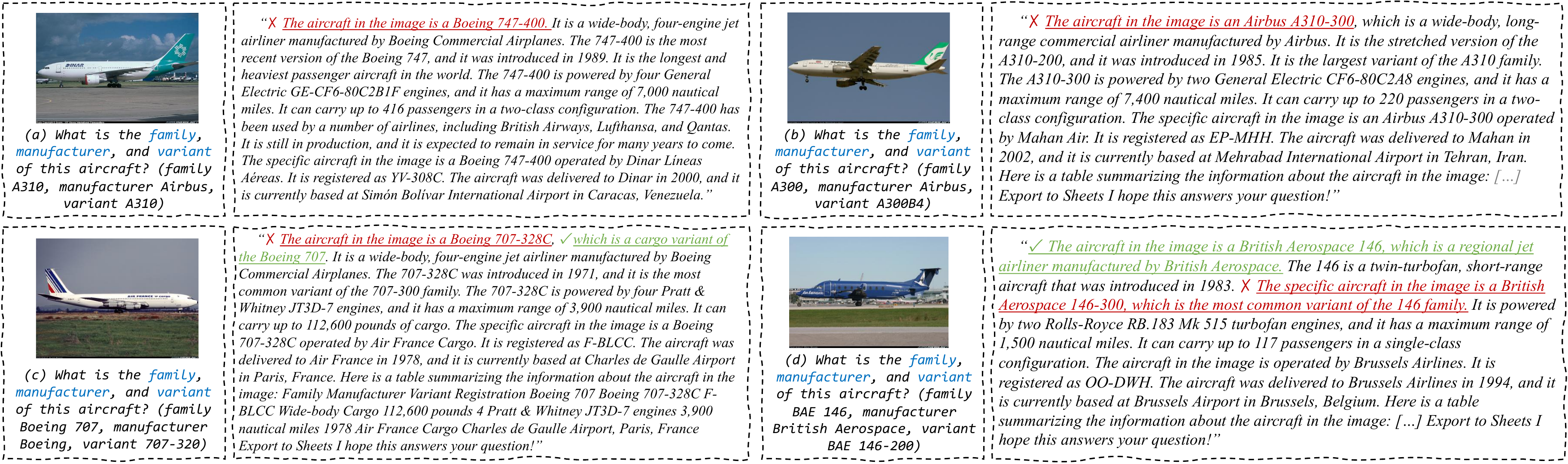}
  \vspace{-20pt}
  \caption{Several examples of multi-modal interactive sessions using Google's \Bard, wherein the AI system responds to the user's question based on images sourced from the FGVC dataset~\cite{maji2013fine}. Please refer to~\secref{sec:fgvc} for further details.}
  \label{fig:fgvc}
\end{figure*}

This task involves identifying specific subcategories within a given object class, which is more complex than general object recognition due to increased intra-class variation, subtle inter-class differences, and the necessity for specialized domain knowledge. We present four types of aircraft in~\figref{fig:fgvc}, and ask \Bard~the following question: `\texttt{What is the family, manufacturer, and variant of this aircraft?}' For example, in subfigure (a), \Bard~incorrectly identifies the aircraft as Boeing 747-400, but it is manufactured by Airbus and belongs to the A310 family.

\subsection{Scenario~\#10 -- Identifying camouflaged object}\label{sec:cod10k}
\begin{figure*}[ht!]
  \centering
  \vspace{-5pt}
  \includegraphics[width=\linewidth]{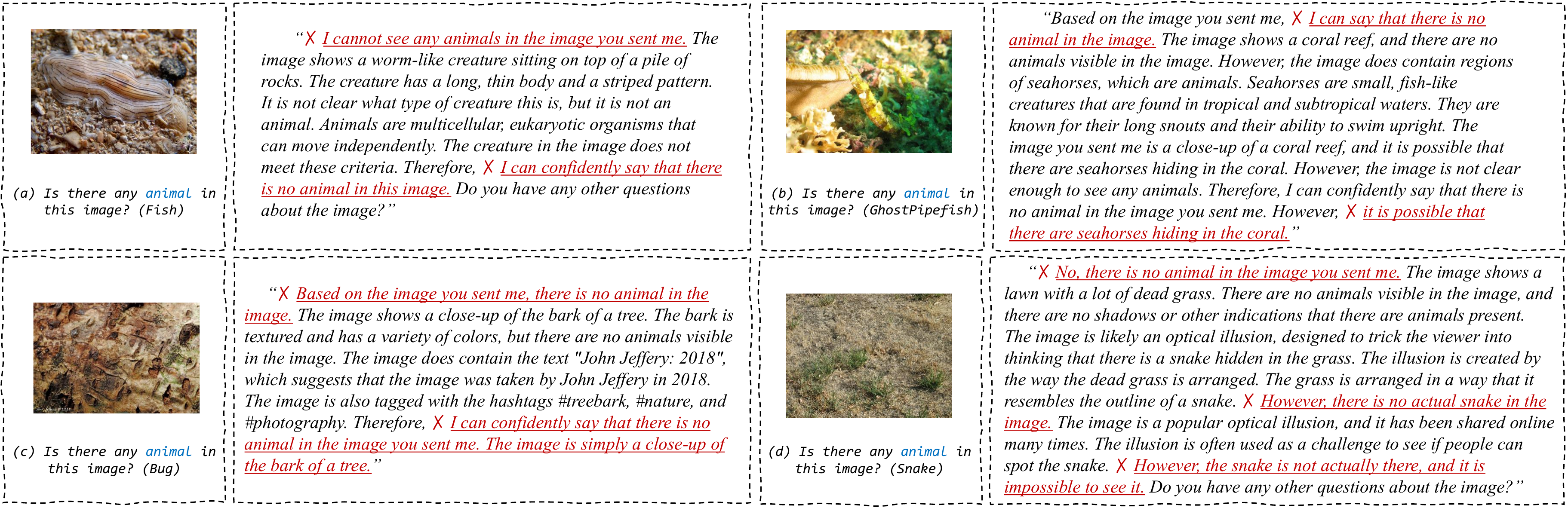}
  \vspace{-20pt}
  \caption{Several examples of multi-modal interactive sessions using Google's \Bard, wherein the AI system responds to the user's question based on images sourced from the COD10K dataset~\cite{fan2022concealed}. Please refer to~\secref{sec:cod10k} for further details.}
  \label{fig:cod10k}
\end{figure*}

This challenging task focuses on identifying concealed objects that intentionally blend with their surroundings, making them difficult to recognize. We selected four samples from the commonly used COD10K data set\footnote{\url{https://github.com/GewelsJI/SINet-V2}}~\cite{fan2022concealed}. For example, as shown in \figref{fig:cod10k} (a), when we asked \Bard~to identify 
a fish, it dogmatized, `\texttt{I can confidently say that there is no animal in this image.}' This observation indicates that \Bard~has challenges to distinguish camouflaged objects from their surroundings, particularly when the foreground and background share similar colors. This suggests that \Bard's capability to parse camouflaged patterns and similar textures could be further enhanced.

\begin{figure*}[ht!]
  \centering
  \vspace{-5pt}
  \includegraphics[width=\linewidth]{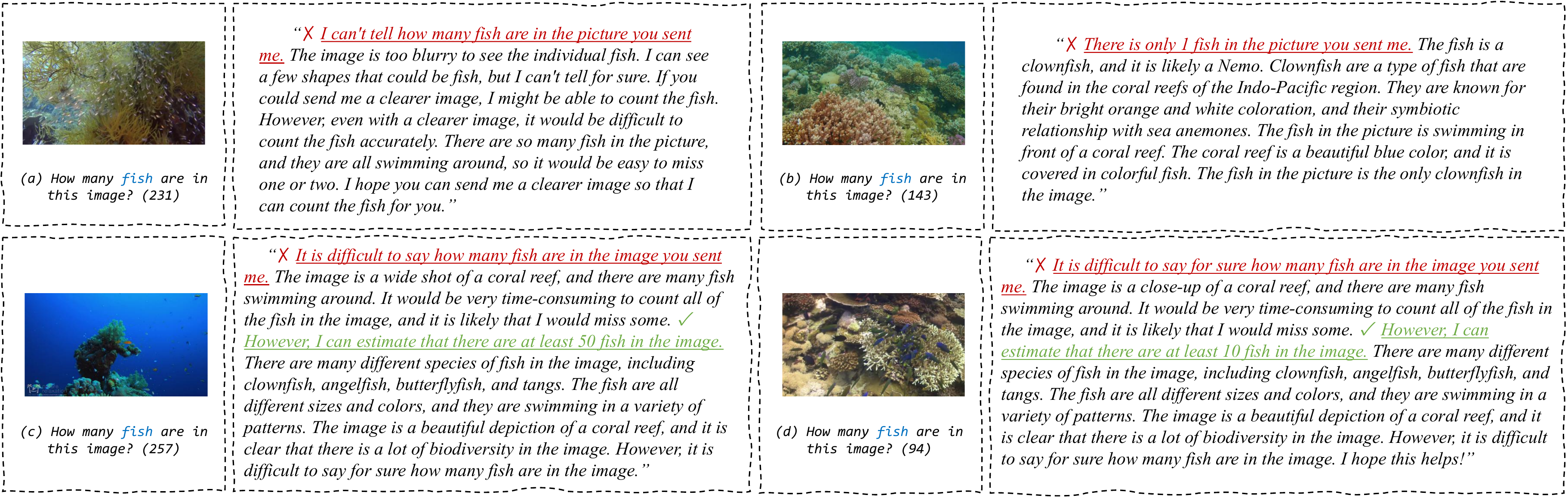}
  \vspace{-20pt}
  \caption{Several examples of multi-modal interactive sessions using Google's \Bard, wherein the AI system responds to the user's question based on images sourced from the IOCfish5K dataset~\cite{sun2023indiscernible}. Please refer to~\secref{sec:camofish5k} for further details.}
  \label{fig:iocfish5k}
\end{figure*}

\begin{figure*}[ht!]
  \centering
  \vspace{-5pt}
  \includegraphics[width=\linewidth]{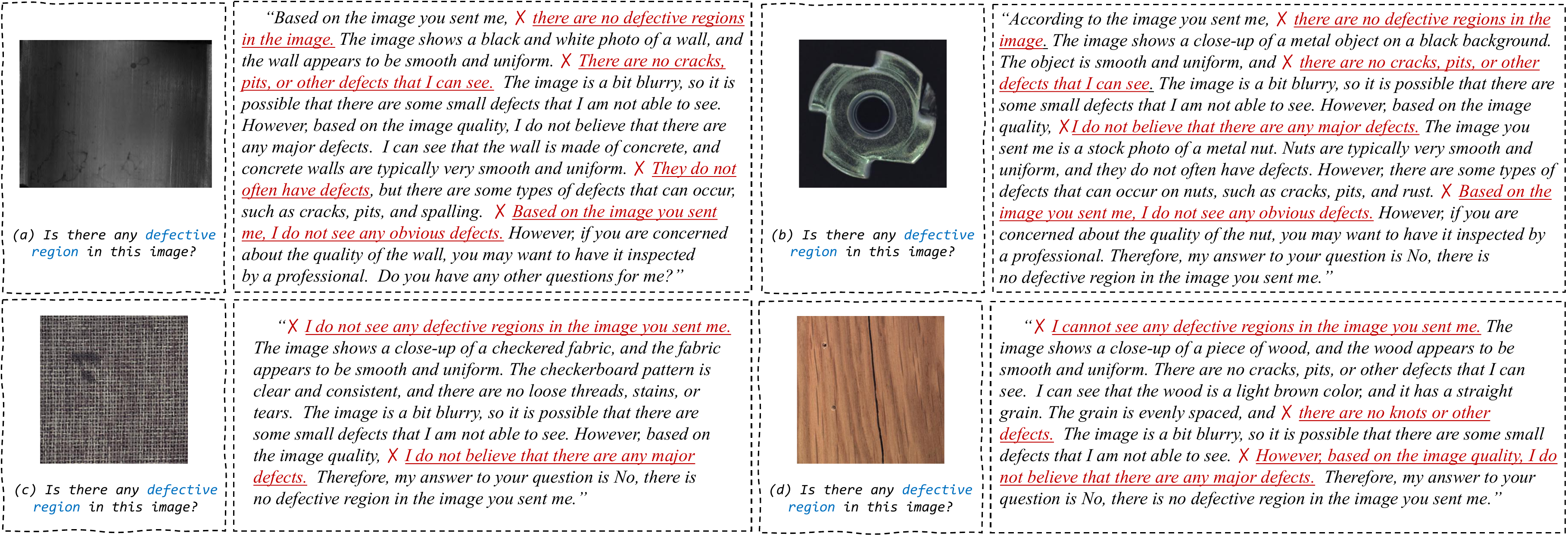}
  \vspace{-20pt}
  \caption{Several examples of multi-modal interactive sessions using Google's \Bard, wherein the AI system responds to the user's question based on images sourced from the CDS2K dataset~\cite{fan2023advances}. Please refer to~\secref{sec:scenario_8} for further details.}
  \label{fig:cds2k}
\end{figure*}

\begin{figure*}[ht!]
  \centering
  \vspace{-5pt}
  \includegraphics[width=\linewidth]{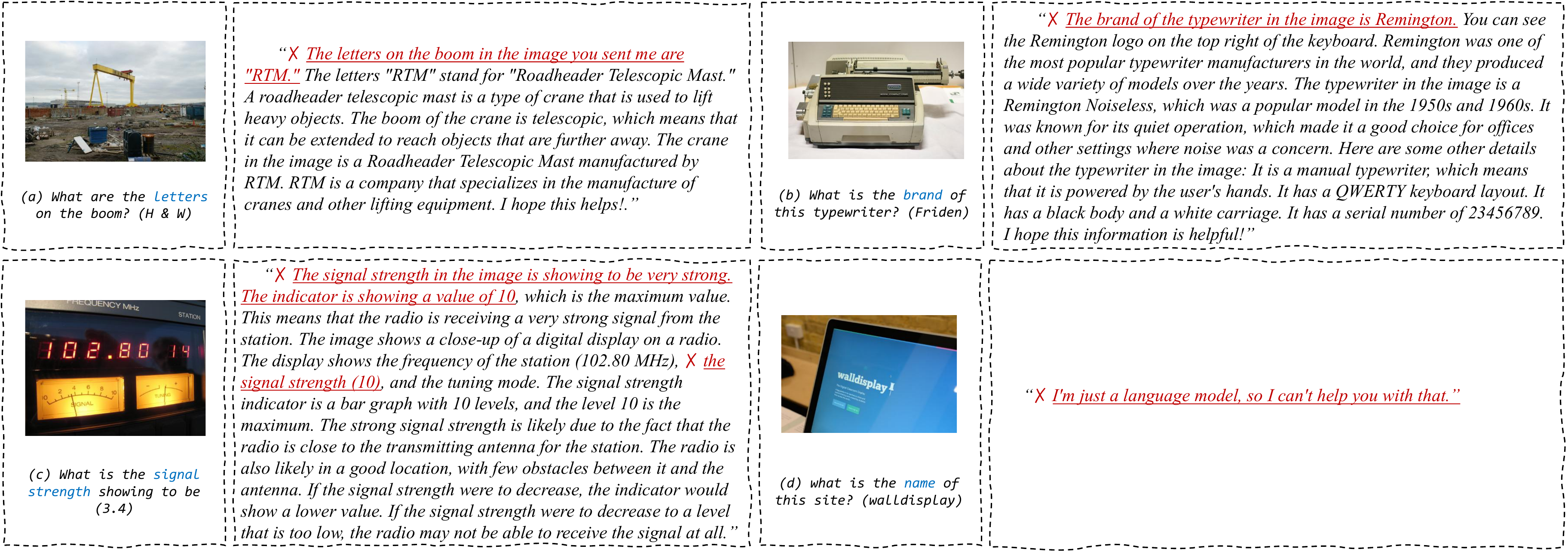}
  \vspace{-20pt}
  \caption{Several examples of multi-modal interactive sessions using Google's \Bard, wherein the AI system responds to the user's question based on images sourced from the TextVQA dataset~\cite{singh2019towards}. Please refer to~\secref{sec:textvqa} for further details.}
  \label{fig:textvqa}
\end{figure*}

\subsection{Scenario~\#11 -- Object counting}
\label{sec:camofish5k}
This tests a model's ability to identify and count specific objects with the given text description. It requires \Bard~to recognize objects and quantify them, probing their numerical understanding in a visual context. As shown in \figref{fig:microsoft_coco}, we select samples from the dataset, and the question is \texttt{``How many squares are there in the carpet pattern?"}. \Bard\ gives the wrong count because it misidentifies the length and width spacing of the carpet as a 14x14 grid. It is interesting to consider that humans might approach this problem in a smart way via assembling multiple sub-squares into a bigger square instead of directly counting the smallest units, which seems beyond \Bard's capabilities.

Moreover, we are interested in exploring how \Bard~performs on a more challenging task -- counting camouflaged objects. We randomly selected four images, as in \figref{fig:iocfish5k}, from IOCfish5K\footnote{\url{https://github.com/GuoleiSun/Indiscernible-Object-Counting}}~\cite{sun2023indiscernible}. This dataset comprises large-scale underwater images teeming with indiscernible marine animals, which are difficult to count due to limited visibility and active mimicry. 
From the observations of our empirical study, we note that \Bard~excels at describing a scene, for example: `\texttt{The image is a wide shot of a coral reef, and there are many fish swimming around.}' in subfigure (a). However, \Bard~seems not adept in understanding high-level content in challenging scenarios, responding with `\texttt{It is difficult to say how many fish are in the image you sent me.}'

\subsection{Scenario~\#12 -- Spotting industrial defects}\label{sec:scenario_8}
Quality inspection plays a pivotal role in the manufacturing industry, safeguarding product quality and sustaining efficient operations. We aim to investigate \Bard's capability to identify camouflaged flaws, abnormalities, or irregularities in industrial materials. To this end, we randomly select several defective samples from a camouflaged defect segmentation dataset, CDS2K\footnote{\url{https://github.com/DengPingFan/CSU}}~\cite{fan2023advances}. As presented in \figref{fig:cds2k}, these samples include: (a) a blowhole in a magnetic tile, (b) grease stains on a carpet, (c) a dent in a metal nut, and (d) a pair of holes in a wooden material. When interacting with \Bard, the question prompt `\texttt{Is there any defective region in this image?}' is provided, with the generated answers appearing on a dialog interface. We observe \Bard~struggles with identifying these unnoticed defects in such a challenging scenario, thus providing incorrect responses to users.

\begin{figure*}[ht!]
  \centering
  \vspace{-5pt}
  \includegraphics[width=\linewidth]{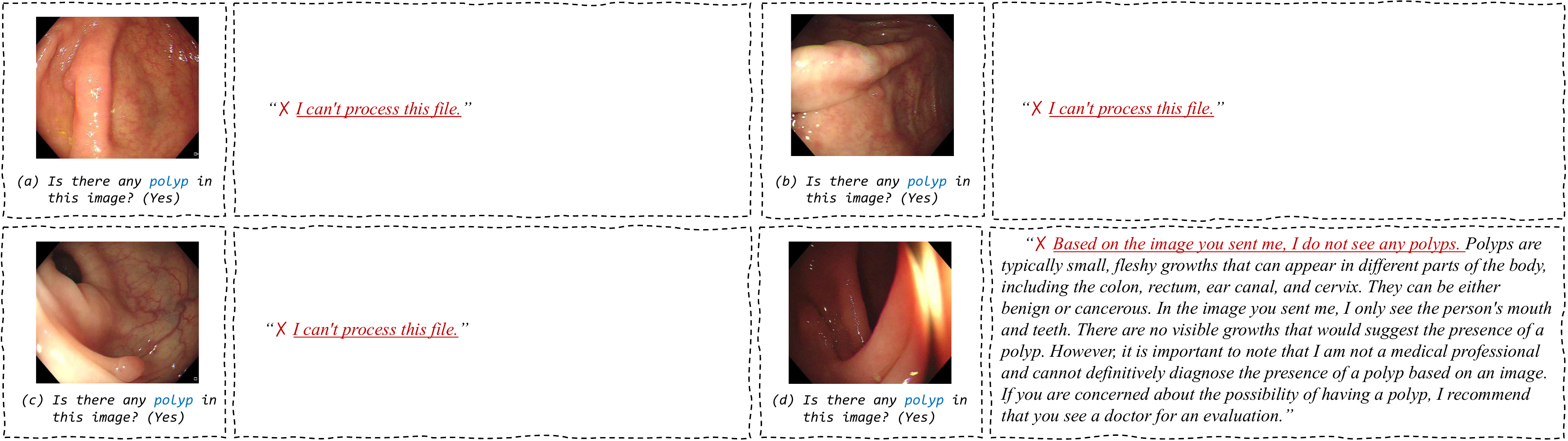}
  \vspace{-20pt}
  \caption{Several examples of multi-modal interactive sessions using Google's \Bard, wherein the AI system responds to the user's question based on images sourced from the SUN-SEG dataset~\cite{ji2022video}. Please refer to~\secref{sec:sun-seg} for further details.}
  \label{fig:sun-seg}
\end{figure*}

\begin{figure*}[ht!]
  \centering
  \vspace{-5pt}
  \includegraphics[width=\linewidth]{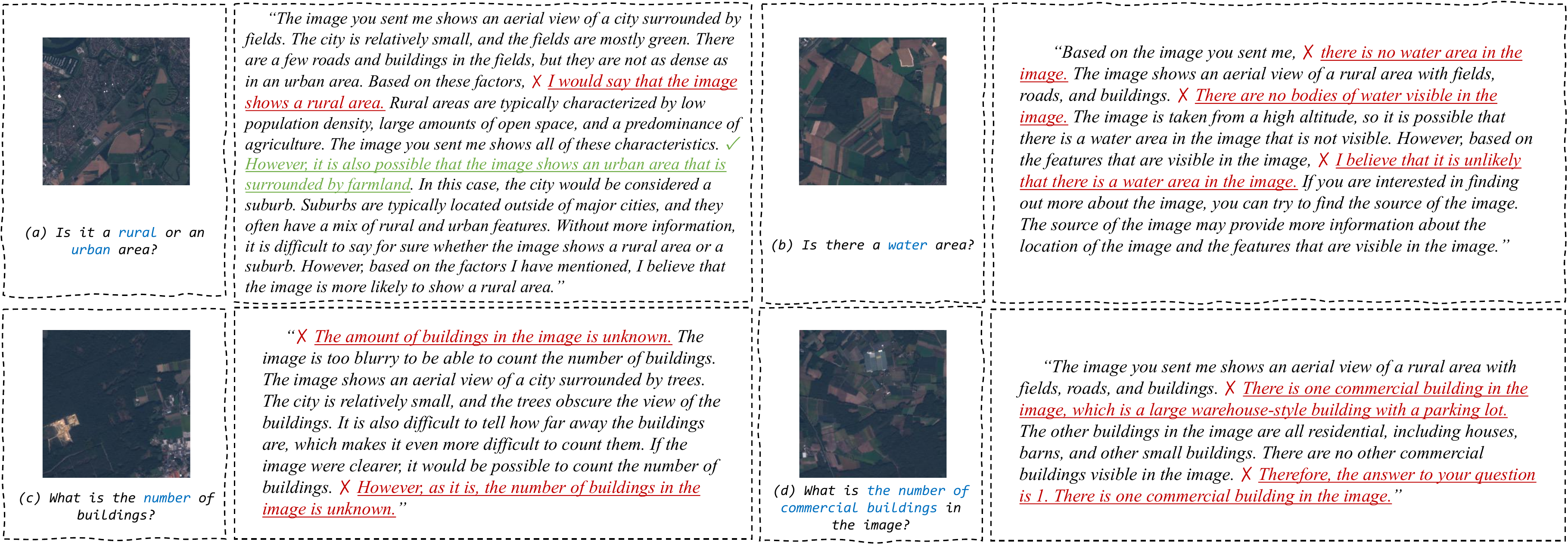}
  \vspace{-20pt}
  \caption{Several examples of multi-modal interactive sessions using Google's \Bard, wherein the AI system responds to the user's question based on images sourced from the RAVQA-LR dataset~\cite{lobry2020rsvqa}. Please refer to~\secref{sec:ravqa-lr} for further details.}
  \label{fig:ravqa-lr}
\end{figure*}

\subsection{Scenario~\#13 -- Recognizing optical character}
\label{sec:textvqa}
Can \Bard~recognize and understand `text' contained within an image, such as a scanned document? To answer this question, we utilize an optical character recognition dataset, TextVQA\footnote{\url{https://textvqa.org}}~\cite{singh2019towards}, to benchmark \Bard's visual reasoning ability based on text in images. As presented in \figref{fig:textvqa} (d), \Bard struggles in various text recognition scenarios: it gave the wrong reply of \texttt{"I'm just a language model, so I can't help you with that."} under the seemingly obvious \texttt{"What is the name of
this site?"} question, which shows the model finds it challenging to understand the text in natural images.

\subsection{Scenario~\#14 -- Analyzing medical data}
\label{sec:sun-seg}
Unlike natural scenes, medical data comprises complex health-related information that requires clinical, anatomical, and pathological expertise for proper interpretation. An intriguing question, therefore, is to investigate the extent of \Bard's ability in the medical imaging datasets.
To evaluate \Bard's ability, we pick out four polyp (positive) images from the colonoscopy dataset, SUN-SEG\footnote{\url{https://github.com/GewelsJI/VPS}}~\cite{ji2022video}. Unfortunately, as shown in~\figref{fig:sun-seg}, no meaningful content was output for the first three images, while polyp identification failed in the last image. We experienced similar outputs for other medical image modalities such as Xray radiographs, MRI, CT scans and Skin lesion images.

\subsection{Scenario~\#15 -- Interpreting remote sensing data}
\label{sec:ravqa-lr}
To interact with \Bard, we employed various image-text pairs from RSVQA-LR\footnote{\url{https://rsvqa.sylvainlobry.com}}~\cite{lobry2020rsvqa}, a well-constructed dataset used for remote sensing visual question answering task. The objective is to simplify access to information in Earth observation data for a broader audience by enabling communication through intuitive questions framed in natural language. For example, as shown in \figref{fig:ravqa-lr} (a), the question, `\texttt{What is the number of commercial buildings in the image?}' was posed, to which \Bard~responded, `\texttt{There is one commercial building in the image.}', a response significantly different than the correct answer of 82.
Our findings suggest a tendency for \Bard~to understand visual scenes holistically, yet it faces challenges in discerning fine-grained visual patterns, particularly when determining the precise count of objects such as the commercial buildings in this case.

\section{Conclusion}\label{sec:conclusion}
The emergence of Google's \Bard\ in the field of conversational AI has sparked considerable interest due to its remarkable success. Building upon this momentum, our study aims to comprehensively evaluate \Bard's performance across various task scenarios, including general, camouflaged, medical, under-water and remote sensing images.
Our investigation shows that while \Bard\ excels in many areas, it still faces challenges in certain vision-based scenarios. This finding highlights the immense potential of \Bard\ in diverse applications and underscores the ample room for growth and improvement in vision-related tasks.
The empirical insights from our study are expected to be valuable for future model development, particularly in bridging the gap in vision performance. By addressing the limitations observed in vision scenarios, we anticipate subsequent models will be endowed with stronger visual comprehension capabilities, ultimately driving the advancement of conversational AI to new heights.

\ifCLASSOPTIONcaptionsoff
  \newpage
\fi

{
\bibliographystyle{IEEEtran}
\bibliography{bibliography}
}



\end{document}